\def\year{2022}\relax
\documentclass[letterpaper]{article} 
\usepackage{aaai22}  
\usepackage{amsfonts}
\usepackage{times}  
\usepackage{helvet}  
\usepackage{courier}  
\usepackage[hyphens]{url}  
\usepackage{graphicx} 
\usepackage{natbib}  
\usepackage{caption} 
\usepackage{pdfpages}

\DeclareCaptionStyle{ruled}{labelfont=normalfont,labelsep=colon,strut=off} 

\usepackage{algorithm}
\usepackage{algpseudocode} 
\usepackage{array}
\usepackage{wrapfig}
\usepackage{float}
\usepackage{multirow}
\usepackage{multicol}
\usepackage{arydshln}
\usepackage{comment}
\graphicspath{ {./images/} }

\usepackage{newfloat}
\usepackage{listings}
\floatstyle{ruled}
\newfloat{listing}{tb}{lst}{}
\floatname{listing}{Listing}
\begin{document}

\title{LSTMSPLIT: Effective SPLIT Learning based LSTM \\on Sequential Time-Series Data}
\author{Lianlian Jiang$^{\star,\dagger}$, Yuexuan Wang$^{\ddagger}$, Wenyi Zheng$^{\ddagger}$, Chao Jin$^{\dagger}$, Zengxiang Li$^{\pm}$, Sin G. Teo$^{\star,\dagger}$\\
$^{\dagger}$\{jiang\_lianlian, jin\_chao, teosg\}@i2r.astar.edu.sg, Institute for Infocomm Research, A*STAR \\ $^{\ddagger}$\{E0509849, E0509848\}@u.nus.edu, National University of Singapore\\$^{\pm}$zengxiang\_li@outlook.com, ENNEW Digital Research Institute, ENN Group, China.
}
\maketitle
\begin{abstract}
\begin{quote}
Federated learning (FL) and split learning (SL) are the two popular distributed machine learning (ML) approaches that provide some data privacy protection mechanisms. In the time-series classification problem, many researchers typically use  1D convolutional neural networks (1DCNNs) based on the SL approach with a  single client to reduce the computational overhead at the client-side while still preserving data  privacy. Another method, recurrent neural network (RNN), is utilized on sequentially partitioned data where segments of multiple-segment sequential data are distributed across various clients. However, to the best of our knowledge, it is still not much work done in SL with long short-term memory (LSTM) network, even the LSTM network is practically effective in processing time-series data. In this work, we propose a new approach, LSTMSPLIT, that uses SL architecture with an LSTM network to classify time-series data with multiple clients. The differential privacy (DP) is applied to solve the data privacy leakage. The proposed method, LSTMSPLIT, has achieved better or reasonable accuracy compared to the Split-1DCNN method using the electrocardiogram dataset and the human activity recognition dataset. Furthermore, the proposed method, LSTMSPLIT, can also achieve good accuracy after applying differential privacy to preserve the user privacy of the cut layer of the LSTMSPLIT.

\end{quote}
\end{abstract}
\noindent 
\section{Introduction}
Machine learning algorithms have been applied to solve many problems during recent years, such as detecting abnormal behavior from normal ones, authorizing access via human facial recognition, enabling robots to learn and do manual tasks by humans, etc. One of the algorithms is deep learning, which has been shown successfully in computer vision, natural language processing, robotics, etc. Traditionally, the data from a single party is used to train and build a deep learning model. It is possible that the deep learning algorithm needs a significant amount of data to prepare a good model. It could limit the model performance as compared to a model trained by various party data sources. However, many parties are reluctant to share their data due to privacy regulations (e.g., GDPA, CCPA and PPDA) and business-sensitive information, especially in the healthcare \cite{Vepakomma2018} and financial industry \cite{Zheng2020, Long2021}.

In recent years, federated learning (FL) \cite{Yang2019} and split learning (SL) \cite{Vepakomma2018, Thapa2021} have been proposed to jointly train models using various data sources from the multi-parties while preserving data privacy. FL enables client sides who have their private and sensitive data to train the models without disclosing their data collaboratively. However, it incurs a high computational cost at the client-side when each party dataset is large. In contrast, SL needs a partial network to run by the client side, and the remaining is run at the server side with high-performance computational resources. Therefore, SL allows various clients to collaborate to train the model with the server. The SL can help amortize the computational burden to the server with the constrained computational resources of the clients while still preserving data privacy. Many solutions use 1D convolutional neural networks (1DCNNs) to tackle the time series classification problem \cite{Cui2016, Abuadbba2020}. Even in \cite{Abuadbba2020}, authors used split architecture. However, the recurrent neural network (RNN) (e.g., LSTM) is practically effective in processing time-series data. To the best of our knowledge, there is still not much work done in SL-based LSTM for several reasons. First, the SL-based LSTM approach is hard to train sequentially partitioned data where the segments of multiple-segment sequential data are distributed across various clients. Another reason is that some existing works \cite{Abuadbba2020} have proven that the approach would cause data privacy leakage in the 1D dataset. The above reasons have motivated us to propose a new direction, LSTMSPLIT, which uses SL with an LSTM network to solve the issues.

Our proposed approach, LSTMSPLIT, consists of the clients and the server. The clients who hold their respective data are willing to jointly train a model where each party only learns its own data but no other party data. The server can be hosted at the public cloud (e.g., AWS, Azure). The approach, LSTMSPLIT, is based on a curious but honest model where the clients and server strictly follow the protocol but could infer information from the output of the LSTMSPLIT. To solve the issue of data privacy leakage in SL that uses 1D datasets, our proposed LSTMSPLIT uses differential privacy (DP) \cite{Cynthia2014}  to add noise to the output of the parties. It helps to preserve data privacy while still achieving a similar performance of the SL-based approaches without DP and of the non-privacy cases. We summarize the contributions of this paper as follows.

\begin{itemize}
\item Proposed and implemented the LSTM network with SL architecture for multiple clients.
\item Verified the effectiveness of the proposed architecture using two datasets.
\item Compared performance of LSTMSPLIT with Split-1DCNN for classification of time series data.
\item Implemented and evaluated the LSTMSPLIT with differential privacy under different levels of privacy.
\end{itemize}

The following Section~\ref{sec:relatedwork} discusses the related work of the SL and federated learning in various applications. The rest of the paper is organized as follows. Sections ~\ref{sec:background} and ~\ref{sec:LSTMsplitmethod} will discuss the background of the Federating Learning, SL and differential privacy, and the proposed method, LSTMSPLIT, respectively. Lastly, the experiment and conclusions and the future work are given in Sections~\ref{sec:experiment} and~\ref{sec:conclusions}, respectively.

\section{Related Work}\label{sec:relatedwork}
FL was proposed by google to train ML models on distributed smart devices without sharing local data to other clients ~\cite{Andrew2018}. FedAvg \cite{Brendan2017} is one of the classic methods for FL strategy. The FedAvg scheme works as follows: Each client gets the same initialized model from the server. The model is then trained at each distributed client with their own data in the first round. Once each client completes its local training, the updated weights are sent back to the central server. Subsequently, the global model is updated at the server side by averaging all the weights received from each client. After that, the server sends the updated global model to each client again for training in the next round. This process repeats until the global model reach its convergence. This method is one of the classic methods for the FL strategy. However, it has the disadvantage of requiring high computational resources at client side when the model becomes complex. For the case where the computational resources at client side is limited, the conventional FL structure is not suitable.

Unlike FL network, SL can split partial network training tasks to the server side. The variants of SL network have been proposed to tackle different dataset types such as time sequence data processing using 1DCNN in ECG signal classification \cite{Abuadbba2020,Kiranyaz2016}, image data processing in health care using 2D convolutional neural network (2DCNN) models \cite{Yadav2019}, etc. In \cite{Vepakomma2018}, a split neural network is proposed for health entities to collaboratively train deep learning models without sharing sensitive raw data. Several configurations of split neural network have been evaluated and the result shows that the split network can provide a higher accuracy than that by the conventional FL method and that by a large batch synchronous stochastic gradient descent method. In \cite{Singh2019}, the analysis results suggest that the SL architect becomes more communication efficient with increasing number of clients and it is highly scalable with the number of model parameters. Whereas, the FL architecture only becomes efficient when the number of data samples is small (1$\sim$4000 clients) or model size is small (1M$\sim$6M parameters). In \cite{Gao2020}, authors evaluated the performance over Internet of Things (IoT)-enabled distributed systems constituted by resource-constrained devices. The results show that FL technique is efficient only when the communication traffic is the first concern. FL performs better than SL in that case because it has a significantly lower communication overhead compared with SL when the number of clients are small. It also demonstrated that neither FL nor SL can be applied to a heavy model with more millions of parameters. In \cite{Abuadbba2020}, authors have implemented the vertical SL architecture for classifying ECG signals using 1DCNN network. It is observed that the 1DCNN model under SL architecture can achieve the same accuracy of 98.9\% like the non-split model. However, it shows that SL may fail to protect the raw data privacy on 1DCNN models. To solve the problem, authors proposed two methods such as 1) adding more hidden layers to the client side 2) applying differential privacy to mitigate the privacy leakage problem. The results show that these two methods are helpful in reducing privacy leakage but they can reduce the accuracy significantly. Therefore, SL structure alone would not be sufficient to maintain the confidentiality of the raw data with 1DCNN models. Instead of working on horizontally or vertically partitioned data only, in \cite{Abedi2021}, authors proposed a Federated SL (FedSL) architecture using RNN to work for the sequentially partitioned data where multiple segments of sequential data are distributed across clients. Based on the result from the simulation and real-world datasets, it demonstrates that the proposed method can train models on distributed sequential data while preserving privacy. It outperforms the centralized FL approach with higher accuracy and fewer communication rounds. However, there is no much work on the scenario where the full time sequence input is hosted at each client side. LSTM model has been widely applied to time series data. As an example, in \cite{Ozal2018}, authors use a bidirectional LSTM model with centralized structure to classify the ECG signals. Instead of directly puting the ECG signals into the LSTM network, authors decomposed the ECG signals into frequency sub-bands at different scales and then used it as sequences for the input of the LSTM network. The result shows an high accuracy of 99.39\%. However, it is a normal centralized structure. Even though many works have been done using 1DCNN or 2DCNN models for classifying ECG signals and using SL architecture in 1DCNN for solving problems as mentioned above, to the best of our knowledge, no much work is found to use LSTM with SL structure.
Since LSTM technique is quite popular and efficient in processing time series data and SL has the advantages of being able to reduce the computational burden from the client side, it is possible to examine the potential capability of the LSTM network with SL architecture. In this paper, we explore the feasibility and effectiveness of the proposed LSTMSPLIT scheme by applying it to two datasets: electrocardiogram (ECG) and human activity recognition (HAR). 
Regarding the privacy leakage, authors in \cite{Vepakomma2020} proposed a 'Nopeek' scheme to preserve the data privacy which is based on reduction of distance correlation between raw data and learned representations during training and inference with image datasets. In \cite{Vepakomma2018_2}, authors reviewed on the distributed deep learning models for training or inference without accessing raw data from clients. 

\section{Background}\label{sec:background}
\subsection{LSTM}
\begin{figure}[h]
\includegraphics[scale=0.22]{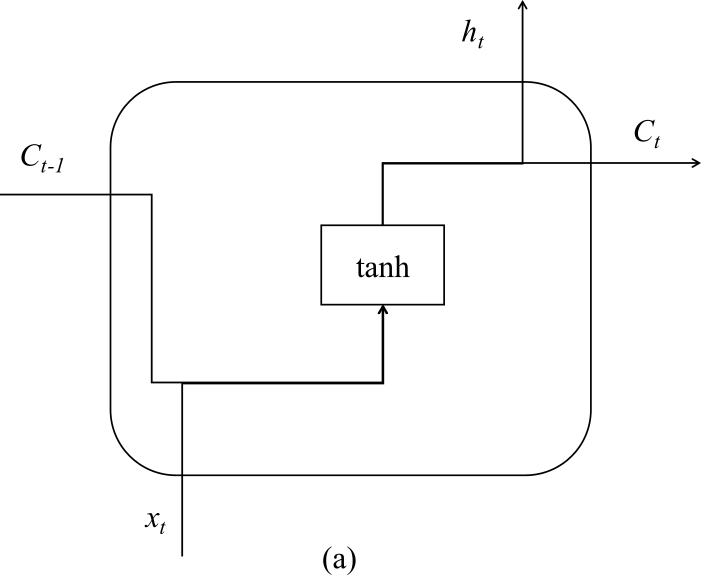}
\includegraphics[scale=0.22]{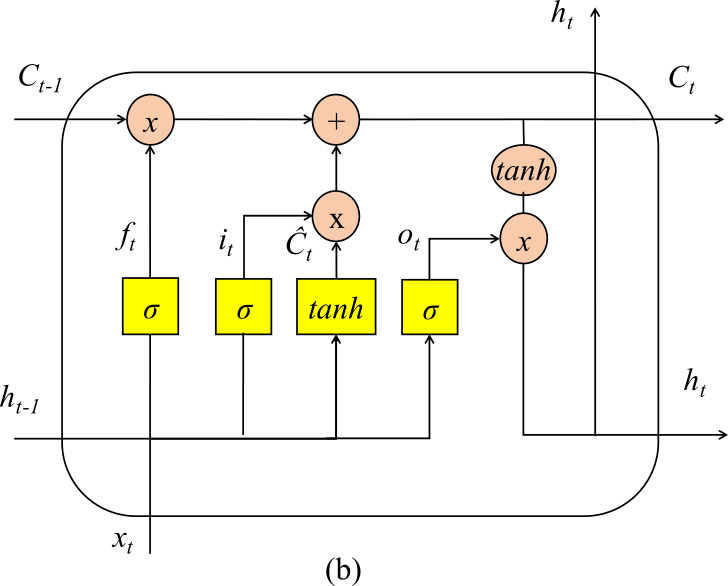}
\caption{Typical internal structure of (a) RNN cell and (b) LSTM cell.}
\label{fig:RNNlstm}
\end{figure}

LSTM neural network is an improved type of RNN. LSTMs were developed to deal with the vanishing and exploding gradient problem encountered when training traditional RNNs \cite{Sepp1997}. Unlike standard feedforward neural networks, LSTM has feedback connections which enable it to not only process the point data without relations to the points in previous time steps, but also sequential data by taking into consideration of points in previous steps. It can selectively remember patterns for a long duration of time. The typical unit structure for RNN and LSTM are shown in Figure \ref{fig:RNNlstm} (a) and Figure \ref{fig:RNNlstm} (b), respectively. Compared to RNN unit which contains only one ``$\mathit{tanh()}$'' function, LSTM cell consists three more gates, namely an input gate, an output gate and a forget gate. The cell remembers values over arbitrary time intervals and the three gates regulate the flow of information into and out of the cell.

\noindent The LSTM controls the information output flow through the cell state by three gates: a forget gate, an input gate, and an output gate. They are composed out of a sigmoid $\/tanh()$ neural net layer ($\sigma$), and pointwise multiplication operations. The cell remembers values over arbitrary time intervals. It is well-known for processing time series data due to its capability to handle lags of unknown duration between important events in a time sequence. 

\noindent The compact formula for LSTM with three gates are given as follows:

\begin{equation}
    i_t=\sigma \left ( W_{ii}x_t+b_{ii}+W_{hi}h_{(t-1)}+b_{hi} \right ),
\end{equation}
\begin{equation}
    f_t=\sigma \left ( W_{if}x_t+b_{if}+W_{hf}h_{(t-1)}+b_{hf} \right ),
\end{equation}
\begin{equation}
    \tilde{C}_t=tanh\left ( W_{ig}x_t +b_{ig}+W_{hg}h_{(t-1)}+b_{hg}\right ),
\end{equation}
\begin{equation}
    o_t = \sigma \left ( W_{io}x_t+b_{io}+W_{ho}h_{(t-1)}+b_{ho} \right ),
\end{equation}
\begin{equation}
    c_t = f_t*c_{(t-1)}+i_t*\tilde{C}_t,
\end{equation}
\begin{equation}
    h_t = o_t*tanh(c_t),
\end{equation}

\noindent where $i_t$ is the activation vector of the input gate. $f_t$ is the activation vector of the forget gate. $x_t$ is the input vector to the LSTM cell. $\tilde{C}_t$ is the activation vector of the input and hidden state. $o_t$ is the activation vector of the output vector. $c_t$ is the cell state vector which is updated by adding regulated results from the forget gate and the input gate. $h_t$ is the hidden state vector which is also known as output vector of the LSTM cell. $W_{ii}$ is the weight matrix between the input neurons and hidden layer. $W_{hi}$ is the weight matrix between the hidden states in the last step and the hidden neurons in the input gate. $'*'$ means the pointwise multiplication. 

\subsection{Split Learning}
\begin{figure}[h]
\includegraphics[scale=0.5]{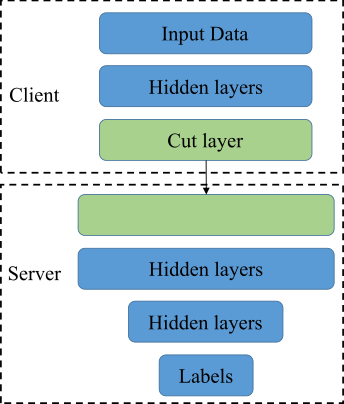}
\centering
\caption{A typical neural network with SL structure.}
\label{fig:SLstructure}
\end{figure}

\noindent A typical SL architecture include the client and server parts as shown in Figure \ref{fig:SLstructure}. The input layer and partial network are processed at the client side, and the rest part of the network is at the server side. Instead of sharing the entire model and weights with all entities, the only communication payloads in the SL are the transformed version of the raw data at the intermediary deep learning layer (also called Cut Layer). 

There are also many other different types of configurations
for SL \cite{Vepakomma2018}, such as the U-shaped SL without label sharing, SL with vertically partitioned data, extended vanilla SL, SL for multi-task output with vertically partitioned input, “Tor”
like multi-hop SL \cite{Roger2004,Vepakomma2018}. The effectiveness of all these architectures need to be explored further. However, this work focuses on the simple vanilla SL architecture where clients do not share input data with the server but server has the access to the labels of the dataset. As the SL architecture limits the calculation at the client side to first few LSTM layers, it can reduce the computation burden at the client side comparing to the centralized learning scheme. The clients do not share its raw input data with server.

\subsection{Differential Privacy}\label{sec_dp}
Differential privacy \cite{Cynthia2014} is one of the most adopted privacy-preserving technologies. It has been rigorously proven to protect user data privacy by adding randomized noise to it. The formal definition of the differential privacy (DP) is given as follows.

{\bf{Differential Privacy:}} A randomized algorithm $\mathbb{A}:\rightarrow \mathbb{R}$ is
$(\epsilon, \delta)$-differential privacy if for the neighboring datasets $D, D^{\prime}$ differing by one element and for all events $S$ in the output space of $\mathbb{A}$ to meet the following condition:

\begin{equation}\label{eq:DP}
Pr[\mathbb{A}(D) \in S] \leq  e^{\epsilon} Pr[\mathbb{A}(D^{\prime}) \in S] + \epsilon.
\end{equation}

\noindent Obviously, $\mathbb{A}$ is $\epsilon$-differential privacy when $\delta$ equals $0$. Another important concept in DP is sensitivity, that measures the maximum difference between the outputs of a pair of the neighboring datasets on a given function $q$ by the following definition.

{\bf{$\ell_2$-Sensitivity:}}
Given a function $q:\mathbb{D}^{T} \rightarrow \mathbb{R}$,
the $\ell_2$-Sensitivity is measured as follows.
\begin{equation}
\triangle (q) =  \max_{D,D^{\prime}} ||q(D) - q(D^{\prime})||,
\label{eq:dp-sensitivity}
\end{equation}

\noindent where $D$ and $D^{\prime}$ are a pair of the neighboring datasets differing by a single element.
From the Equation~\ref{eq:dp-sensitivity}, it indicates that the larger the sensitivity of the function $q$, the much easier for an adversary to get information in the dataset. The issue can be solved by adding sufficient noise to the function $q$ so as to defend inference and construction attacks from the adversary. In other words, the user privacy of the dataset is well protected.

\section{LSTMSPLIT: Our Practical and Secure Collaborative Method} \label{sec:LSTMsplitmethod}

\subsection{The LSTMSPLIT Approach}
\begin{figure}[h]
\includegraphics[scale=0.35]{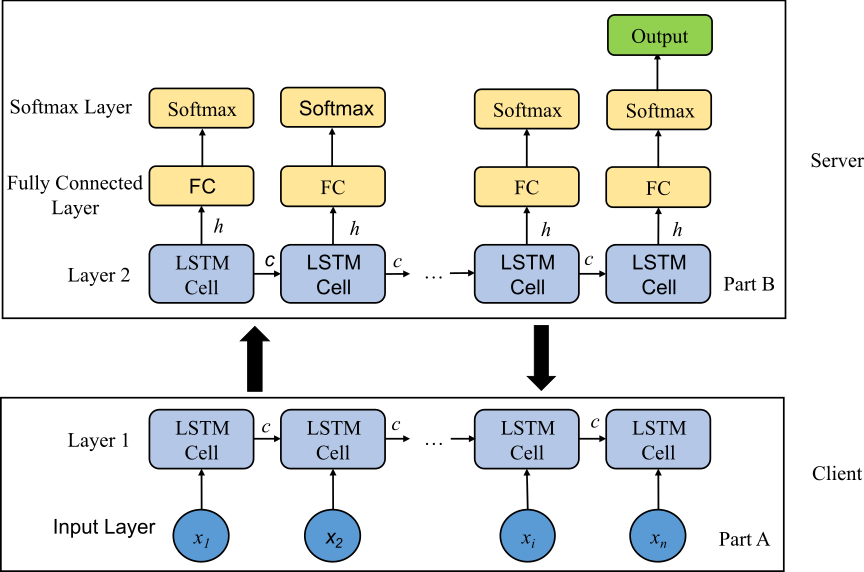}
\centering
\caption{SL architecture of the LSTM network.}
\label{fig:SL+LSTM}
\end{figure}

\noindent In the SL architecture, partial components are running at the server or client-side. As shown in Figure~\ref{fig:SL+LSTM}, the multilayer LSTM model is split into the client-side and server-side. After initializing the weights at both sides of server and client, whenever there is a new training request, the client then carries out forward propagation with the new dataset and sends the activation outputs of its hidden states at the cut layer and labels to the server. Once the server receives the output from the client, it calculates forward propagation. The forward activation function with the gradient information is passed between the client and server sides to train a joint model collaboratively. After obtaining the loss function, the server runs the backpropagation and sends the gradients of the loss function w.r.t. the activations of the hidden states back to the client. When the client receives the gradients of the cut layer, it back-propagates the gradients received from the server and updates the weights at its own side.

This SL is hard to apply to the LSTM architecture with a single layer. Therefore, when using the LSTM network in the time-series data, the entire length of the input sequence is usually stored at one client-side. Thus, we split the LSTM from $c^{th}$ layer (also called a cut layer) instead of splitting the network based on the input steps. The labels of the datasets together with the activation functions at the client side are sent to the server-side.

\subsubsection{SL with Multiple Clients}
\begin{figure}[h]
\includegraphics[scale=0.28]{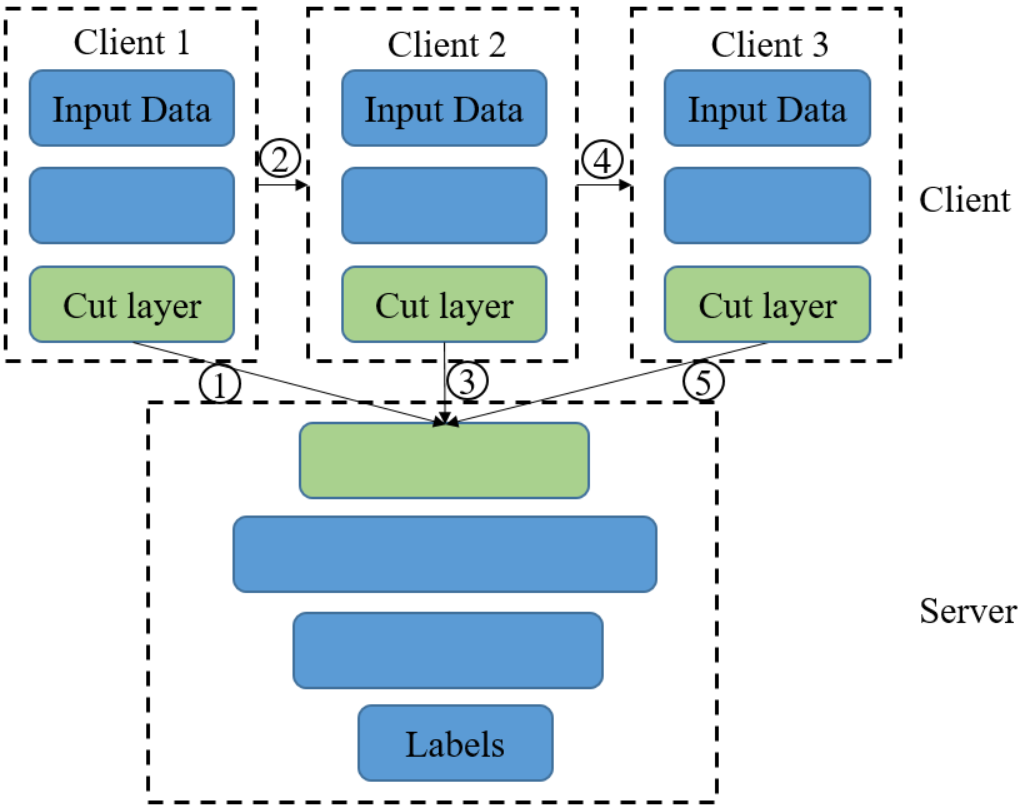}
\centering
\caption{An example of SL training structure with multiple clients.}
\label{fig:SLmultiClients}
\end{figure}

Figure \ref{fig:SLmultiClients} shows an example of the split learning architecture for the LSTM network with multiple clients. The training process starts from Client 1, which trains the LSTM network with a server in Step 1 and passes the trained weights to Client 2, as shown in Step 2. Client 2 then continues to train the network by collaborating with the server with its own dataset in Step 3. Once it completes its training, it sends the updated weights to Client 3 in Step 4. When Clients 3 receives the signal, it continues to train the network with the server in Step 5. This process can be triggered whenever there is a new training request. This configuration is suitable for a multi-modal multi-institutional collaboration. Clients with data from a specific domain can collaborate with each other to train a partial model up to the cut layer. A new client who has a new set of domain data can join the training process to improve the accuracy of final results. Our proposed method, LSTMSPLIT, is based on the
SL architecture, as shown in Figure \ref{fig:SLmultiClients}.

\noindent The detailed steps for implementing our proposed LSTMSPLIT at a client-side and server-side are given in Algorithm 1 and Algorithm 2, respectively.

\subsubsection{LSTMSPLIT Client}
\begin{algorithm}[t!]
	\caption{LSTMSPLIT Client} 
	\begin{algorithmic}[1]
	\Require Training data  $D_{train}^{c_i}$ of the client $c_{i \in \left \{1, \ldots, k\right \}}$. 
    \State Initialize weight $ W^{c_i} \leftarrow \O $.
	\For{each client $c_{i\in 1,\ldots, k}$ }
	    \State Set $W^{c_{i+1}}\leftarrow W^{c_{i}} $ 
	    \For{each epoch, $e_{j\in 1,\ldots, m}$}
	       \State Feedforward propagation $W^{c_i}_{e_j}$ with $D_{train}^{c_i}$.
	       \State Calculate activation function $A_{e_j}^{c_i}$ at its cut layer.
	       \State Send $A_{e_j}^{c_i}$ with their labels $Y^{c_i}$ to the server $s$.
	       \State Receive gradient $d_{A^{c_i}_{e_j}}^{s}:= \bigtriangledown \mathit{l}\left ( A^{c_i}_{e_j};W^{c_i}_{e_j} \right )$
	       \State Backward propagation with $d_{A^{c_i}_{e_j}}^{s}$.
	      \State Update the weight $W^{c_{i}}_{e_{j+1}}= W^{c_{i}}_{e_{j}}-\eta d_{A^{c_i}_{e_j}}^{\prime s}$.
	 \EndFor
	 \EndFor
	\end{algorithmic} 
\end{algorithm}
\noindent Suppose that a LSTM network with $N$ hidden layers is split between $c_{th}$ layer and $(c+1)^{th}$ layer, as shown in Figure~\ref{fig:SL+LSTM}. In fact, the proposed LSTMSPLIT client can use two different configuration modes to update the weights of the network \cite{Gupta2018,Thapa2021}: a centralized mode or peer-to-peer mode. 
In the centralized mode, the client uploads weights to either the server in the system or a third-party server. When there is a new request from a client to train the network, it downloads the weights from the server. In contrast, in the peer-to-peer mode, the server sends the address of the client last trained to the current training client. The current client updates its client-side model by directly connecting to the address of the last trained client and then downloading the latest trained weights.

After the information of the last trained client is requested and received by a client, it updates its own weights with the weights of the previous client. The client then carries out feedforward calculation in each epoch and sends the activation output to the server-side. Subsequently, the client waits for the gradients to be sent from the server side. Once the gradients at the cut layer are received from the server-side, the client continues to back-propagate until all the weights at the client-side are updated. This process is repeated until all the training epochs are completed. The updated weights of the client are then passed into the next client to continue the same training process. 

\subsubsection{LSTMSPLIT Server}
\noindent Similar to the client-side, the network structure of the LSTMSPLIT Server is shown in Figure~\ref{fig:SLmultiClients}.
After obtaining the output data from a client, the server carries out a feedforward calculation. The loss and the gradients are then calculated until the cut layer. Subsequently, the gradients are sent back to the client. After that, the train losses of data are accumulated and the current prediction is measured. Finally, the accuracy of training can be computed in each epoch. 

\begin{algorithm}[t!]
	\caption{LSTMSPLIT Server} 
	\begin{algorithmic}[1]
    \Require The activation function $A^{c_i}$ of the cut layer and the labels $Y^{c_i}$  of the client $c_{i}$. 
    \State Set weight $W^s = W^{\prime s}$. 
    \For{each epoch, $e_{j\in 1,\ldots, m}$}
        \State Feedforward propagation $W^{s}$ with $A^{c_i}$.
        \State Calculate loss with labels $Y^{c_i}$ and predictions $\tilde{Y^{c_i}}$.
        \State Backpropagration.
        \State  $W_{e_j+1}^s \leftarrow W_{e_j}^{s}-\eta \bigtriangledown l\left ( A_{e_j}^s,W_{e_j}^s \right )$.
        \State Send $d_{A^{c_i}_{e_j}}^{s}=\bigtriangledown l\left (A_{e_j}^s,W_{e_j}^s \right )$ to client $c_i$.
        
    \EndFor
    \State Update weight $ W^{\prime s} =  W^{\prime s} +  W^s$. 
	\end{algorithmic} 
\end{algorithm}

\section{Experiment}\label{sec:experiment}
\noindent In this section, we evaluate the performance of the proposed method, LSTMSPLIT, with two datasets. We first discuss the two datasets and the settings for our experiment. Then, we show the experiment results and discuss its performances. Finally, we further discuss how to protect the privacy leakage at the cut layer of the proposed method, LSTMSPLIT, using differential privacy.

\subsection{Dataset}
\subsubsection{ECG Data}
\noindent 
ECG dataset is extracted from the MIT-BIH arrhythmia dataset \cite{ECGdataset2001}. Similar to \cite{Wu2018,Abuadbba2020}, we collected 26,490 samples in total which contains five different heartbeat types, namely $N$ (normal beat), $L$ (left bundle branch block), $R$ (right bundle branch block), $A$ (atrial premature contraction), and $V$ (ventricular premature contraction). 
The filtered samples as in \cite{Abuadbba2020} are used to feed into the Split-1DCNN and LSTMSPLIT. The details of the ECG dataset are shown in Table \ref{table:ECGdata}. 

\begin{table*}[t]
\centering
\caption{The specification of the ECG dataset.}
\begin{tabular}{ c|m{6em}| m{6em}| m{6em}| m{8em}| m{6em}|c } 
 \hline
 {\bf Dataset size } & {\bf Normal beat (N)} &{\bf Left bundle  branch block (L)} &{\bf Right bundle branch block (R)}  & {\bf Atrial premature contraction (A)} & {\bf Ventricular premature contraction (V)} &{\bf Total} \\ \hline
 Total & 6000 & 6000 & 6000 & 2490 & 6000 & 26490\\ 
 \hline
\end{tabular}
\label{table:ECGdata}
\end{table*}

\subsubsection{HAR Data}
\noindent HAR dataset contains six types of human activities, walking ($W$), walking upstairs ($WU$), walking downstairs ($WD$), sitting ($S$), standing ($SD$), and laying ($L$). The human activities of 30 volunteers who wore a smartphone on the waist were collected \cite{HARdata}.
With the accelerometer and gyroscope embedded, 3-axial linear acceleration and 3-axial angular velocity data at a constant rate of 50Hz were captured. The original signals are pre-processed and the time and frequency domain features were calculated to form a 
561-dimensional vector as an input to the Split-1DCNN and LSTMSPLIT networks.
The details of the HAR datasets are shown in Table \ref{table:HARdataset}. 

\begin{table*}[h]
\centering
\caption{The specification of the HAR dataset.}
\begin{tabular}{ c|c|m{5em}|m{5em}|c|c|c|c} 
 \hline
 {\bf Dataset size} & {\bf Walking (W)} & {\bf Walking Upstairs (WU)} & {\bf Walking Downstairs (WD)}  & {\bf Sitting (S)} & {\bf Standing (SD)} & {\bf Laying (L)} & {\bf Total }\\ \hline
 Total & 1722 & 1544 & 1406 & 1777 & 1905 & 1944 & 10298 \\ 
 \hline
\end{tabular}
\label{table:HARdataset}
\end{table*}

\subsection{Experiment Setting}
\noindent 
As 1DCNN is one of the popular deep learning methods for processing time-series data,
we implement split learning based on 1DCNN, namely Split-1DCNN, to compare our proposed method, LSTMSPLIT. Both Split-1DCNN and LSTMSPLIT use multiple clients (5, 10, 15, 20, 25, and 30 clients) with the two datasets as discussed before in the experiment. The two datasets are both shuffled and each client randomly selects samples from them.
The server trains the network with one client and moves to another in sequence until all the clients complete their training processes. We can add a scheme of choosing a group of clients to participate in each round. It can be either based on the training performance of previous rounds or randomly select the clients \cite{Thapa2021}.

Each dataset is shuffled and 20\% of the dataset is separated as the testing data for all the clients. All the clients use the same testing data to measure classification accuracy, while the rest 80\% of the dataset is divided and assigned to each individual client. 

Both Split-1DCNN and LSTMSPLIT are implemented in Python 3.8 with Pytorch 1.7. 
The experiments were run in the machine with following specifications: NVIDIA GeForce RTX 2080Ti 11GB CUDA GPU, 64GB RAM, x64-based processor 8-core Intel(R) CPU @ 3.60GHz.

For a fair comparison, the general training parameters of Split-1DCNN and LSTMSPLIT are the same. 
The settings of the LSTM and 1DCNN network architecture and its training parameters are summarized in Table \ref{table:ParaLSTM} and Table \ref{table:Para1DCNN}, respectively.

\begin{table}[h!]
\centering
\caption{Parameter settings for the multilayer LSTM network in SL structure with multiple clients.}
\begin{tabular}{ m{8em}|c|l }
 \hline
{\bf Parameters} & {\bf ECG dataset} & {\bf HAR dataset} \\ \hline
Input series length & 128 & 561 \\\hline
Number of LSTM layers & \multicolumn{2}{c}{2}\\\hline
Number of neurons in hidden layer 1 & \multicolumn{2}{c}{200}\\\hline
Number of neurons in hidden layer 2 & \multicolumn{2}{c}{200}\\\hline
Batch size & \multicolumn{2}{c}{32}\\\hline
Epoch &	\multicolumn{2}{c}{200}\\\hline
Learning rate &	\multicolumn{2}{c}{0.0001}\\\hline
\end{tabular}
\label{table:ParaLSTM}
\end{table}

\begin{table}[h!]
\centering
\caption{Parameter settings for the 1DCNN network in SL structure multiple clients.}
\begin{tabular}{ m{8em}|c|l }
 \hline
{\bf Parameters} & {\bf ECG dataset} & {\bf HAR dataset} \\ \hline
Input series length	& 128 &	561\\ \hline
Number of ‘Conv’ layers 
(convolutional layers + LeakyReLU() + MaxPool1d)& \multicolumn{2}{c}{2}	\\ \hline
Fully connected layer
(linear() + LeakyRelu()) & \multicolumn{2}{c}{2}\\ \hline
Number of neurons in the second hidden layer of fully connected layer &	\multicolumn{2}{c}{128} \\ \hline
Batch size & \multicolumn{2}{c}{32}	\\ \hline
Epoch & \multicolumn{2}{c}{200}\\ \hline
Learning rate& \multicolumn{2}{c}{0.0001}\\ \hline

\hline
\end{tabular}
\label{table:Para1DCNN}
\end{table}

\subsection{Performance Metrics}
The target of both Split-1DCNN and LSTMSPLIT is to correctly predict the category of each sample in both the ECG and HAR datasets. The prediction accuracy is measured by:
\begin{equation}
Accuracy = \frac{n_{c}}{n_{e}}\times 100\%,
\end{equation}
\noindent where $n_{c}$ is the number of classes classified correctly and $n_{e}$ is the size of the testing dataset.

Time complexity (training time) of Split-1DCNN and LSTMSPLIT 
is measured by:
\begin{equation}
TC = K\times E\times B\times t_b,
\end{equation}
\noindent where $K$ is the total number of clients, $E$ is the total number of the epochs, $B$ is the total number of data batches set for each client during the training process and $t_b$ is the training time required for each batch of data. Lastly, the communication complexity is the total amount of time the clients and server send and respond from each other.
\subsection{Discussion}

\begin{table*}[h!]
\centering
\caption{The performance of LSTMSPLIT and Split-1DCNN network on ECG dataset.}
\begin{tabular}{ccccc}
\hline
                                   & \textbf{Clients} & \textbf{Time Complexity (ks)} & \textbf{Comm Complexity (ks)} & \textbf{Test Accuracy (\%)} \\ \hline
\multirow{6}{*}{Split-1DCNN + ECG} & 5       & 1.1885             & 0.0517               & 90.13              \\ \cline{2-5} 
                                   & 10      & 1.1403            & 0.0528               & 89.94              \\ \cline{2-5} 
                                   & 15      & 1.1088             & 0.0514               & 89.91              \\ \cline{2-5} 
                                   & 20      & 1.0822             & 0.0500               & 90.21              \\ \cline{2-5} 
                                   & 25      & 1.0420             & 0.0460               & 91.12              \\ \cline{2-5} 
                                   & 30      & 1.0551             & 0.0488               & 89.94              \\ \hline
\multirow{6}{*}{LSTMSPLIT + ECG}   & 5       & 6.8101             & 0.7869              & 98.50              \\ \cline{2-5} 
                                   & 10      & 5.2806             & 0.9368              & 91.82              \\ \cline{2-5} 
                                   & 15      & 6.2908             & 0.9205              & 96.69              \\ \cline{2-5} 
                                   & 20      & 5.4010             & 0.9359              & 96.44              \\ \cline{2-5} 
                                   & 25      & 6.7603             & 0.7959              & 96.08              \\ \cline{2-5} 
                                   & 30      & 6.1590             & 0.8807              & 94.56              \\ \hline
\end{tabular}
\label{table:ECGperf}
\end{table*}

\begin{table*}[h!]
\centering
\caption{The performance of LSTMSPLIT and Split-1DCNN network on HAR dataset.}
\begin{tabular}{ccccc}
\hline
                                   & \textbf{Clients} & \textbf{Time Complexity (ks)} & \textbf{Comm Complexity (ks)} & \textbf{Test Accuracy (\%)} \\ \hline
\multirow{6}{*}{Split-1DCNN + HAR} & 5       & 0.8712              & 0.0229               & 97.85              \\ \cline{2-5} 
                                   & 10      & 0.4471              & 0.0248               & 97.80              \\ \cline{2-5} 
                                   & 15      & 0.4258              & 0.0249               & 97.46              \\ \cline{2-5} 
                                   & 20      & 0.7914              & 0.0205               & 97.61              \\ \cline{2-5} 
                                   & 25      & 0.3991              & 0.0214               & 97.41              \\ \cline{2-5} 
                                   & 30      & 0.3584              & 0.0192               & 97.66              \\ \hline
\multirow{6}{*}{LSTMSPLIT + HAR}   & 5       & 6.5529             & 1.0983             & 93.36              \\ \cline{2-5} 
                                   & 10      & 7.0486             & 1.2057             & 89.16              \\ \cline{2-5} 
                                   & 15      & 7.0326             & 1.4648             & 89.65              \\ \cline{2-5} 
                                   & 20      & 6.4677             & 1.2286             & 89.01              \\ \cline{2-5} 
                                   & 25      & 7.7693             & 1.8821             & 88.72              \\ \cline{2-5} 
                                   & 30      & 6.1622             & 1.0921             & 88.53              \\ \hline
\end{tabular}
\label{table:HARperf}
\end{table*}

\begin{figure*}[h!]
\includegraphics[scale=0.5]{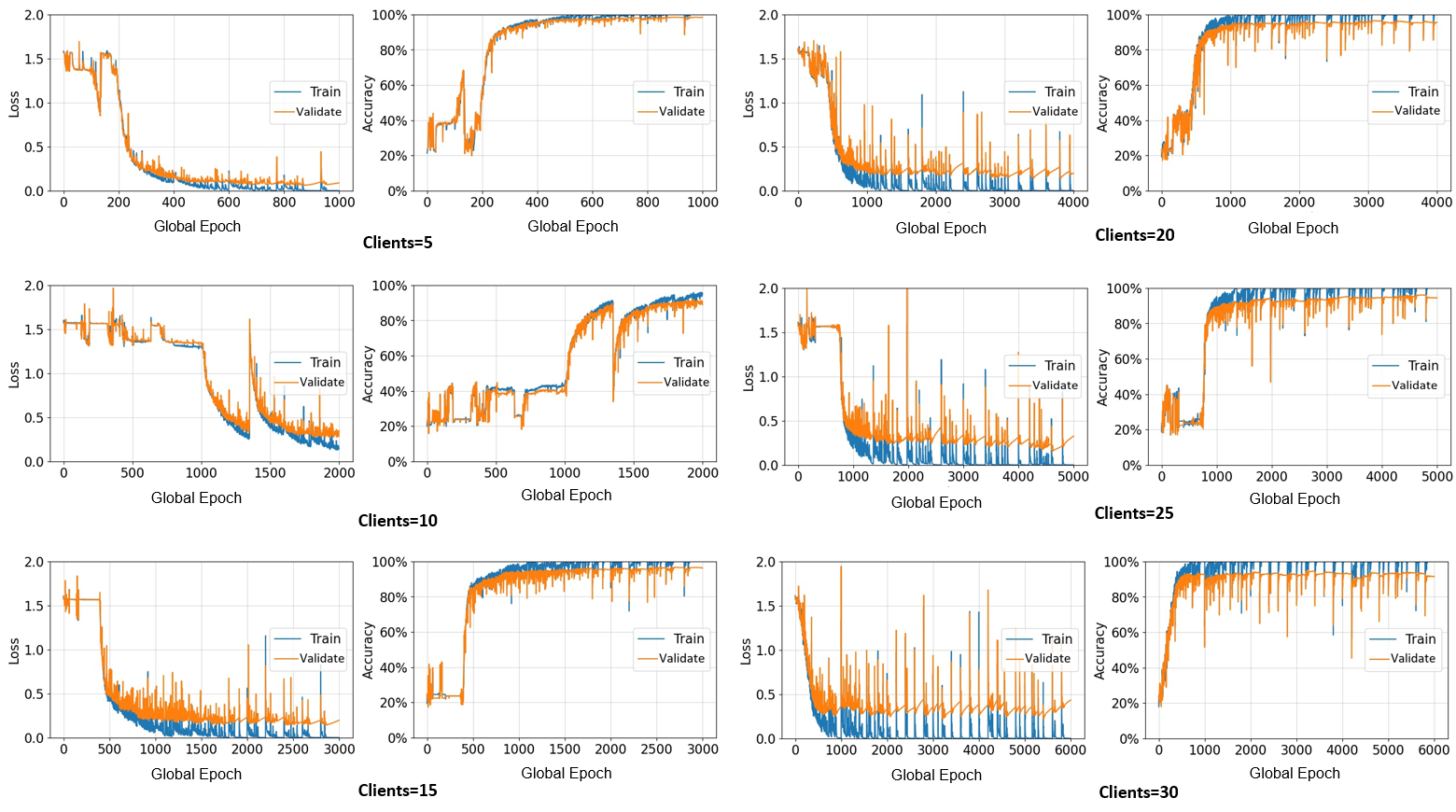}
\centering
\caption{Losses and accuracy during training and testing process of LSTMSPLIT network with ECG dataset.}
\label{fig:Figure12_splitLSTM_ECG2}
\end{figure*}

\begin{figure*}[t!]
\includegraphics[scale=0.5]{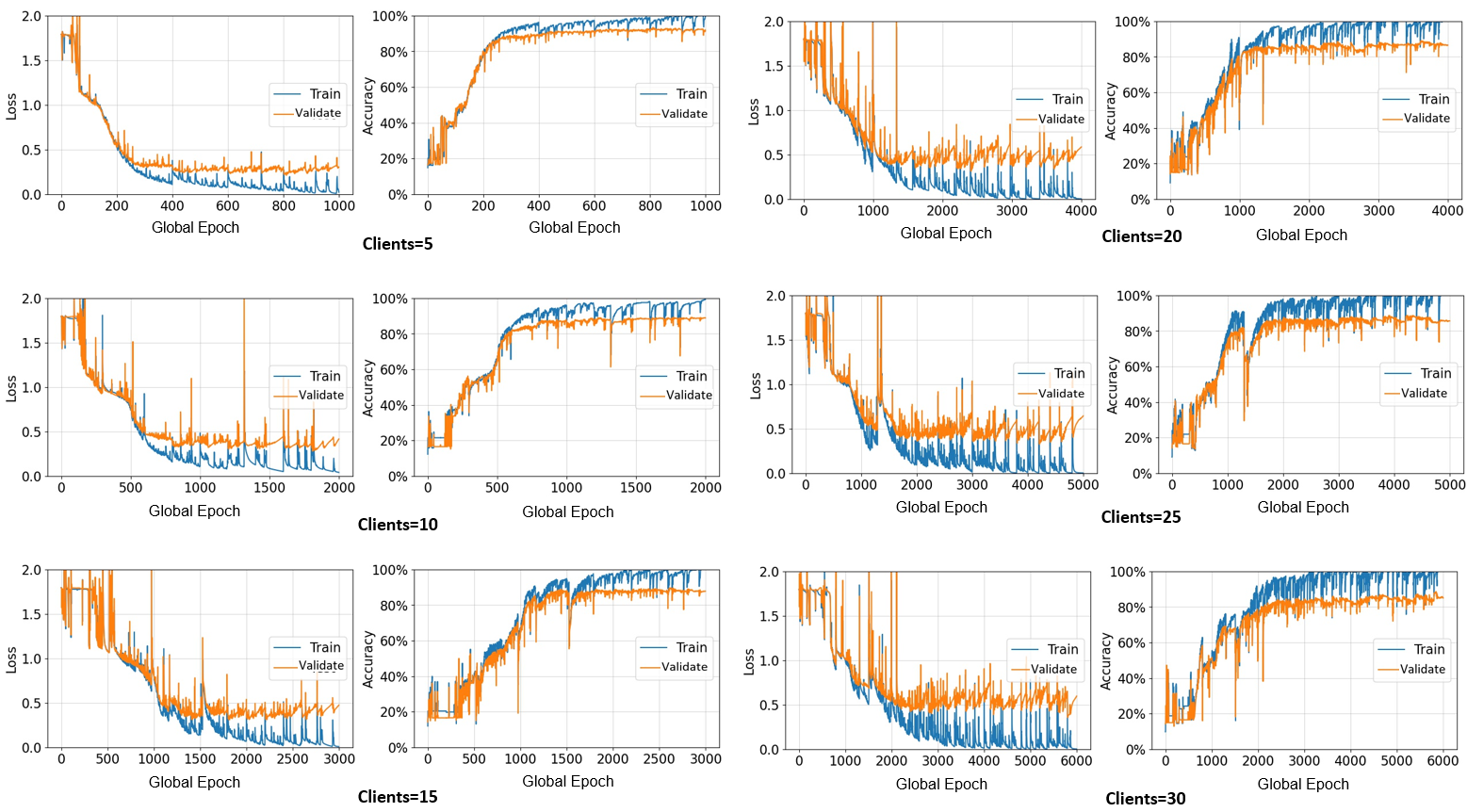}
\centering
\caption{Losses and accuracy during training and testing process of LSTMSPLIT network with HAR dataset.}
\label{fig:Figure13_splitLSTM_HAR2}
\end{figure*}

\subsubsection{Classification for ECG Data}
Table \ref{table:ECGperf} shows the performance of the Split-1DCNN and LSTMSPLIT on ECG dataset. From Table \ref{table:ECGperf}, we can see that the LSTMSPLIT outperforms Split-1DCNN in classifying ECG dataset. The proposed LSTMSPLIT network can give a higher classification accuracy compared to Split-1DCNN for different client settings. For example, for the case of five clients, LSTMSPLIT can give 98.50\% in accuracy, whereas the Split-1DCNN can only give 90.13\% under the similar setting of the training parameters. 
The third column of the Table \ref{table:ECGperf} shows the time complexity of LSTMSPLIT with different clients. We can see that the time complexity for different clients with the same dataset is similar. The reason is that the total number of the training dataset for each case with different numbers of clients are the same. Data are equally divided and assigned to each client. The time complexity of Split-1DCNN with varying clients is also similar as the reason stated above.

From Table \ref{table:ECGperf}, we can also see that a longer training time is required for LSTMSPLIT than the Split-1DCNN on the same number of epochs. This is quite obvious because it has a more complex network architecture in LSTMSPLIT, which has been set with many (200) hidden neurons for each LSTM layer. Less hidden neurons set in LSTM cells can significantly reduce the training time but users should keep in mind that it may degrade the performance of the proposed LSTMSPLIT.
However, users can make the trade-off between the complexity and the accuracy according to their own requirements.

The loss and accuracy the LSTMSPLIT during training for ECG dataset on different clients (5,10, \dots, 30) are shown in Figure~\ref{fig:Figure12_splitLSTM_ECG2}. Each client is trained with 200 epochs. The global training continues until all the clients complete their training process. 
The results show that Split-1DCNN has a faster convergence speed than the LSTMSPLIT with a smoother decrease in the loss values (The change of the loss and accuracy curves for Split-1DCNN are not shown due to the page limit). However, our proposed LSTMSPLIT can reach a higher accuracy even hits high fluctuations during the training process.

The trade-off between the time complexity and accuracy of the LSTMSPLIT is based on different requirements as fewer hidden neurons may affect the prediction accuracy. Typically, the accuracy can also be improved by increasing the number of training epochs. Again, we set 200 epochs in this experiment only to perform training time for catering to different testing scenarios. Since optimizing the training parameters is not the focus of this work, the proposed LSTMSPLIT can be further tuned to improve the prediction accuracy.
\subsubsection{Classification for HAR Data}
\noindent Table \ref{table:HARperf} shows the performance of the LSTMSPLIT and Split-1DCNN on the HAR dataset. From Table \ref{table:HARperf}, we can see that both LSTMSPLIT and Split-1DCNN can reach reasonable and good accuracy. However, in the HAR dataset, LSTMSPLIT does not outperform the Split-1DCNN. Therefore, Split-1DCNN is more suitable in classifying human activity as the network performance also depends on the characteristics of the data. Please note that in this paper, we aim to provide a workable solution of Split Learning based on LSTM to handle sequential time-series data for the classification problem. It can be one of the solutions that uses choose from to solve their problems for time series data. The proposed LSTMSPLIT can always be further tuned with different hyperparameter settings or optimization mechanisms to get higher accuracy.
The losses and accuracy during the training and testing process for LSTMSPLIT on HAR dataset is shown in Figure~\ref{fig:Figure13_splitLSTM_HAR2}.
Again, Split-1DCNN shows a smoother decrease in the loss values during training and testing process compared to LSTMSPLIT.
\subsubsection {LSTMSPLIT with DP}
\noindent To verify the effectiveness of the proposed LSTMSPLIT in further preserving data privacy of the cut-layer with the strategy of DP, we add noise to the cut layers output by varying epsilon-delta values of the DP. Due to the page limitation, the discussion of the results of LSTMSPLIT with DP is skipped in this paper.

\section{Conclusion and Future Work}\label{sec:conclusions}
A Split Learning (SL) architecture based on the LSTM called LSTMSPLIT is proposed. In the LSTMSPLIT structure, multiple layers of the LSTM network are applied. Partial LSTM layers are trained at the client side, and the rest layers are trained at the server-side. 
Multiple clients can jointly train in sequence using the proposed LSTMSPLIT. To further improve the data privacy protection of the cut layer in LSTMSPLIT, differential privacy (DP) is used by adding noise to the output of the cut-layer of each client and the server. This protection strategy helps preserve data privacy while still achieving a similar performance of the LSTMSPLIT without DP. The effectiveness of the proposed LSTMSPLIT has been proven with the experiment using the two datasets. 
The proposed LSTMSPLIT is practically effective in processing sequential time-series data. It can handle time-series data in the split architecture to reduce the computational burden on the client-side while still achieving good performance with user privacy protection. We will investigate how to apply our LSTMSPLIT with different partitioned data types and on non-IID time-series data in future work.

\bibliography{LSTMSPLIT.bib}

\end{document}